\documentclass[runningheads]{llncs}
\usepackage{graphicx}
\usepackage{multirow}
\usepackage{tikz}
\usepackage{comment}
\usepackage{hyperref}
\usepackage{amsmath,amssymb} 
\usepackage{color}
\usepackage{cite}
\usepackage[accsupp]{axessibility}  
\usepackage{booktabs}

\newcommand{\name}{\textit{3DLatNav}}

\begin{document}

\pagestyle{headings}
\mainmatter

\title{3DLatNav: Navigating Generative Latent Spaces for Semantic-Aware 3D Object Manipulation} 

\titlerunning{3DLatNav}
\author{Amaya Dharmasiri \inst{1} \and
Dinithi Dissanayake\inst{1} \and
Mohamed Afham \inst{1} \and Isuru Dissanayake \inst{1} \and Ranga Rodrigo \inst{1} \and Kanchana Thilakarathna \inst{2}}
\authorrunning{Dharmasiri et al.}
%
\institute{Department of Electronic and Telecommunication Engineering, University of Moratuwa, Sri Lanka \and
The University of Sydney}

\maketitle
\vspace{-4mm}
\begin{figure*}[h]
  \centering
  \includegraphics[width=1\linewidth]{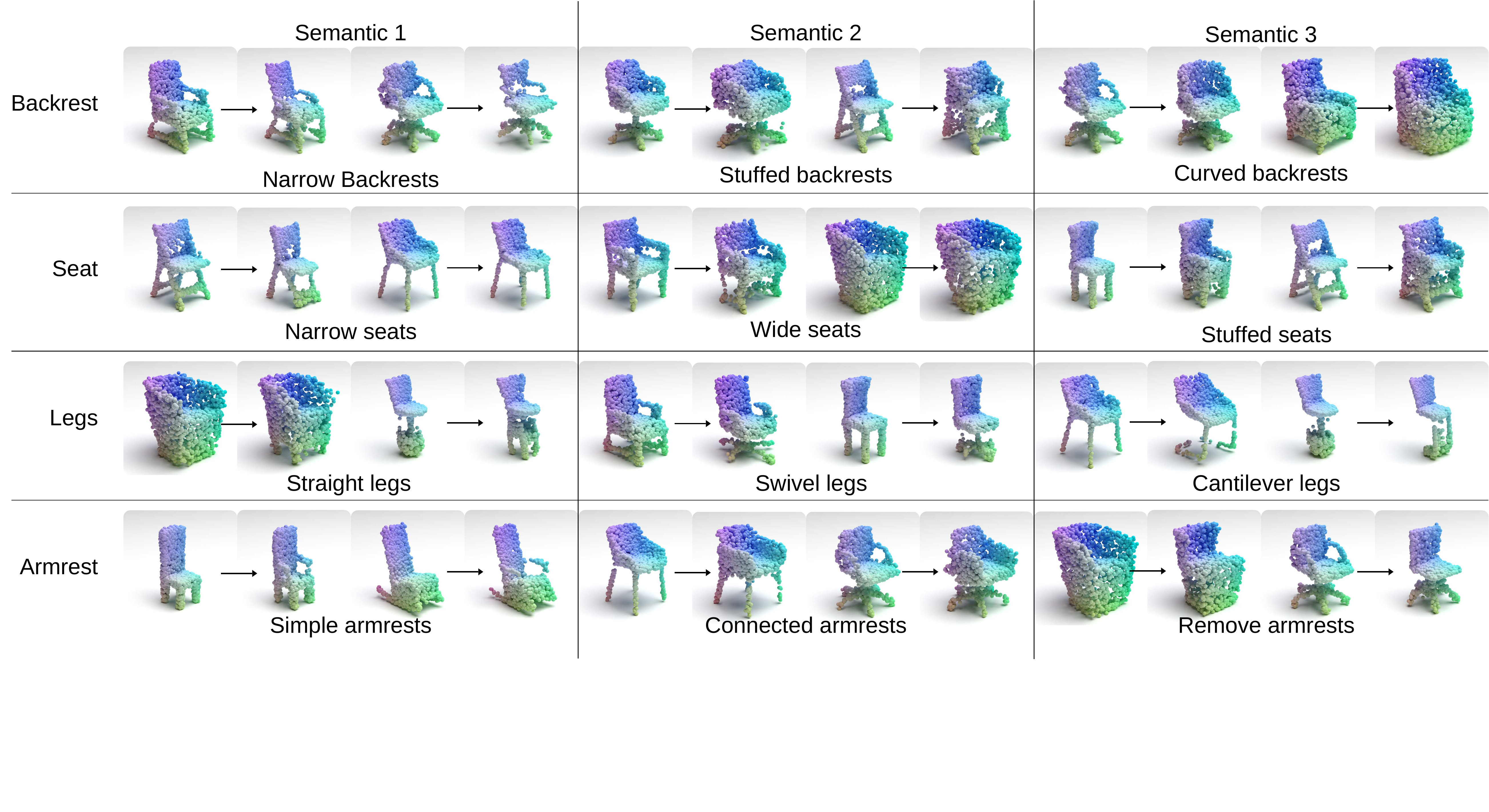}
  \vspace{-8mm}
  \caption{Versatile interpretable latent directions discovered by \(\name\) in 3DAAE\cite{Zamorski20203dAAE} generative latent space. 
  Part semantic attributes of sample objects are manipulated by translating their latent codes to the required latent space direction. The changes are limited to the intended part while the features of other parts are preserved.}
  \label{fig:single_attrib}
  \vspace{-10mm}
\end{figure*}

\begin{abstract}
3D generative models have been recently successful in generating realistic 3D objects in the form of point clouds. 
However, most models do not offer controllability to manipulate the shape semantics of component object parts without extensive semantic attribute labels or other reference point clouds.
Moreover, beyond the ability to perform simple latent vector arithmetic or interpolations, there is a lack of understanding on how part-level semantics of 3D shapes are encoded in their corresponding generative latent spaces.
In this paper, we propose \(\name\); a novel approach 
to navigate pretrained generative latent spaces to enable controlled part-level semantic manipulation of 3D objects.
First, we propose a part-level weakly supervised shape semantics identification mechanism using latent representations of 3D shapes. Then, we transfer that knowledge to a pretrained 3D object generative latent space to unravel disentangled embeddings to represent different shape semantics of component parts of an object in the form of linear subspaces, despite the unavailability of part-level labels during the training.
Finally, we utilize those identified subspaces to show that
controllable 3D object part manipulation can be achieved by applying the proposed framework on any pretrained 3D generative model.
With two novel quantitative metrics to evaluate the consistency and localization accuracy of part-level manipulations, we show that \(\name\) outperforms existing unsupervised latent disentanglement methods in identifying latent directions that encode part-level shape semantics of 3D objects.
With multiple ablation studies and testing on state-of-the-art generative models, we show that \(\name\) can implement controlled part-level semantic manipulations on an input point cloud while preserving other features and the realistic nature of the object.

\keywords{Latent Navigation, Latent Disentanglement, 3D Semantic Manipulation, Controlled 3D regeneration}
\end{abstract}

\section{Introduction}
\label{sec:intro}

The increasing popularity of applications that rely on 3D vision in domains such as virtual or augmented reality, gaming and entertainment leads the way to true realization of metaverse in the near future. 
The advancements of deep learning and the availability of large-scale 3D point cloud datasets have enabled fast progress in the 3D vision tasks such as 
classification~\cite{charles2017pointnet,wang2019dgcnn,Liu2019RelationShapeCN,Thomas2019KPConvFA,Xu2021PAConvPA,Zhao_2021_pointtransformer,Zhang2019ShellNetEP,crosspoint}, 
segmentation~\cite{Zhao_2021_pointtransformer, charles2017pointnet,Thomas2019KPConvFA,Li2018PointCNNCO,Huang2018RecurrentSN,Zhang2019ShellNetEP}, 
3D reconstruction~\cite{groueix2018atlasnet,Kar2017LearningAM,mescheder2019OccupancyNetworks,Xu2019DISNDI}, and
shape and view synthesis~\cite{peng2021neural,Kwon2020RotationallyTemporallyCN,Mildenhall2020NeRFRS,Niemeyer2020DifferentiableVR}. 
Controlled generation or editing of 3D point cloud objects is of growing interest due to its ability to develop interactive 3D vision applications~\cite{cihan2021LPMnet,Jiang2019DisentangledRL,kaichun2020Structedit}. 
3D generative models have been successful in learning overall representations of specific object classes~\cite{Achlioptas2018firstGAN,guandao2020pointflow,Zamorski20203dAAE,Shitong2021diffpm,Riojin2021ShapeGF,Gadelha2018MultiresolutionTN}. 
However, these models cannot enable controllable manipulation of part-level semantics in point clouds, unless fine granular semantic labels and definitions or reference target point clouds are provided.
This significantly limits the utility of existing methods. We hypothesise that this is also due to the lack of understanding and prior work of how semantic features are embedded in 3D generative latent spaces, although there is a large body of work in 2D generative latent spaces~\cite{Shen2020interfaceGAN,alec20162DGAN,bau2019gandissect,Shen2021ClosedFormFO,Hosoya2019GroupbasedLO,Hadad_2018_CVPR,sheila2018MLVAE,Ansari2019HyperpriorIU}. 
(Object parts as refered in the rest of the paper has the obvious meaning. part-Semantics or part-level-shape-semantics are different features/types of the said part. (Eg: Part- legs. Par-semantic- swivel legs)). 


To this end, we raise and address the following questions in this paper; 
\begin{itemize}
    \item How can we define and identify part-level semantic features for 3D point cloud objects when such annotations are unavailable?
    \item How does a latent code of an object embed different parts and different semantic attributes of each component part? 
    \item How can we manipulate shape-semantics of a given point cloud when we do not have a reference point cloud to extract the target features from? 
\end{itemize}

Hence, we formulate \textbf{\(\name\)}: a novel simple framework to part-semantic aware manipulation or generation of objects through \textbf{3D} generative \textbf{Lat}ent space \textbf{Nav}igation. 
\(\name\) can independently manipulate part-semantics of an object. In contrast to prior work, our approach does not require part-level semantic feature annotations, a reference point cloud to extract features of expected end result, or specialized post-processing. 
First, we generate latent representations for independent object parts. (Eg: Backrests)
We implement agglomerative clustering in this latent space to find different shape-semantics within the specific object part. (Eg: Curved, reclined etc. Backrests).  
Next, we utilize the identified part-level semantic categories to find linear subspaces in a pre-trained object-level generative latent space.
As such, we unravel that 3D object generative models learn disentangled representations for the semantics of object parts, despite the unavailability of part-level labels during training. 
Finally, we demonstrate that one or a combination of identified part-semantics can be manipulated on a query object by simply translating its latent vector to a weighted combination of corresponding linear subspaces within the generative latent space. 
(Unlike approaches that swap parts between different objects\cite{cihan2021LPMnet, anastasia2019composite}, the extent of manipulation of an added shape semantic in \(\name\) is controlled by tuning a weight parameter).
We validate the performance of \(\name\) on multiple object classes from ShapeNet dataset~\cite{shapenet2015} with state-of-the-art 3D generative models~\cite{Zamorski20203dAAE, guandao2020pointflow, Shitong2021diffpm}.

The paper presents the following contributions: 
\begin{itemize}
\item We propose a method to extract independent shape semantics from different 3D object parts in a weakly supervised manner 
and prove that they can be used learn disentangled latent directions that encode part-level shape semantics for any pre-trained 3D generative model.  
\item We present comprehensive analysis and ablation studies about how part-level semantics behave and are encoded 3D generative latent spaces.
\item We show that such behaviors can be used to controllably manipulate object parts, hence we introduce \(\name\); a framework to manipulate one or many independent part semantics of 3D point clouds 
in a preferred way, purely by navigating latent spaces of a pre-trained object generators, 
while preserving the object’s quality and realistic nature.
\item We introduce two quantitative metrics to measure the \textit{localization ability} and \textit{consistency} of part semantics controlled/ manipulated by
\(\name\) and show that it outperforms existing unsupervised latent disentanglement methods in identifying meaningful directions.

\end{itemize}

\section{Related Work}
\label{sec:related work}

\subsection{3D Generative models}

\textbf{Point Cloud Generation}: 
There have been plethora of work in 3D point cloud generative models 
These generative frameworks vary from tree structure based graph convolutions \cite{Gadelha2018MultiresolutionTN,Dong2020treeGCN, Valsesia2019LearningLG} to progressive point embeddings \cite{Wen_2021_pointembeddings} and attention based modeling\cite{Sun2020PointGrowAL}. Autoencoders \cite{Zamorski20203dAAE, Yang_2018_folding_net}, flow based models \cite{guandao2020pointflow, Pumarola_2020_Cflow} and energy based models \cite{Xie2021GenerativePD, Shitong2021diffpm} are increasingly becoming popular. 
3D Shape modeling and generations by learning disentangled representations~\cite{tewari2022disentangled3d,zhu2018visual,li2021spgan, Yang2022disen} 
also produce interesting results. \emph{The focus of our paper is to understand the properties of these generative latent spaces and propose methods to introduce model agnostic semantic control for point cloud generation, because none of these existing generative models offer such control without the prior knowledge of the end point cloud. }
\\
\textbf{Part Based Modeling}: Several of the 3D generative models~\cite{Junli2020Pagenet,nadav2019CompoNet} use part-based decomposing techniques to generate new point cloud objects. \cite{anastasia2019composite} uses a decomposer and a composer for part-aware shape embedding and synthesis, permitting for swapping and randomly assembling of the parts. These have several drawbacks centered on the symmetry and the connectivity of the edited 3D objects. \cite{kaichun2019structernet, Roberts_2021_ICCV} use hierarchical aspects to generate new structures. \emph{However, these methods need extensive fine grained semantic labeling and specialised training, whereas we propose a framework that is scalable and applicable to any pre-trained generative model.} 
Instead of generating parts separately, \cite{cihan2021LPMnet} introduces an autoencoder model that modifies and generates 3D objects  with respect to their parts. Such methods require specialized post processing to perform the editing. \emph{In contrast, \(\name\) does not require any post processing beyond the pre-trained generator} 

\subsection{Understanding Generative Latent Spaces}
\textbf{2D Generative Latent Spaces}: Varying input latent codes of generative model, has been explored as a way of manipulating the generated output. The unit based specialization in synthesizing visual concepts \cite{bau2019gandissect} and the vector arithmetic properties~\cite{alec20162DGAN,upchurch2017deep2DGAN}
 of Generative Adversarial Networks(GANs) were explored in 2D domain for controlled generations. More recently, InterfaceGAN \cite{Shen2020interfaceGAN} followed by \cite{Shen2021ClosedFormFO} explains how semantics are encoded in the latent space of 2D GANs, providing a face editing framework. However, these work rely on annotated semantic labels in 2D datasets. Due to the unavailability of such semantic labels and definitions in 3D objects, understanding of 3D generative latent spaces has not been investigated thoroughly in literature. The above frameworks, have not been generalised or tested for adaptability to 3D use cases. 
 \\
\textbf{3D Generative Latent Spaces}: There are a few unsupervised methods for manipulating 3D generative latent spaces. ShapeVAE \cite{Nash2017shapeVAE} measures the activity of each latent unit and demonstrates that high activity latent units tend to encode significant shape changes. \cite{rios2020featurevisuali3D} tries to understand the activations in 3D point cloud AEs, and visualizes the features by clustering in the feature space. These changes or features neither localise into one part of the object nor comprehend the expected manipulation. Latent space interpolation and latent space arithmetic 
 is commonly used by 3D generative and disentanglement methods\cite{Achlioptas2018firstGAN,groueix2018atlasnet,Shitong2021diffpm,Park2019DeepSDFLC,Pumarola_2020_Cflow,Nash2017shapeVAE,hao2019global2local, Zamorski20203dAAE, Rakotosaona2020IntrinsicPC, tristan2019Geodisentan} to manipulate the semantics. 
This also needs a source object and a reference object which has the intended semantic we want to change, to either interpolate or perform latent arithmetic.
\emph{In this paper, using the available part annotations for 3D point cloud objects, we introduce a novel definition for part based semantics. Using these semantic labels we unravel the latent spaces of generative models to enable independently controlling object parts by editing the latent code, without the need of a reference point cloud.} 


\begin{figure*}[t]
  \centering
  \includegraphics[width=0.85\linewidth]{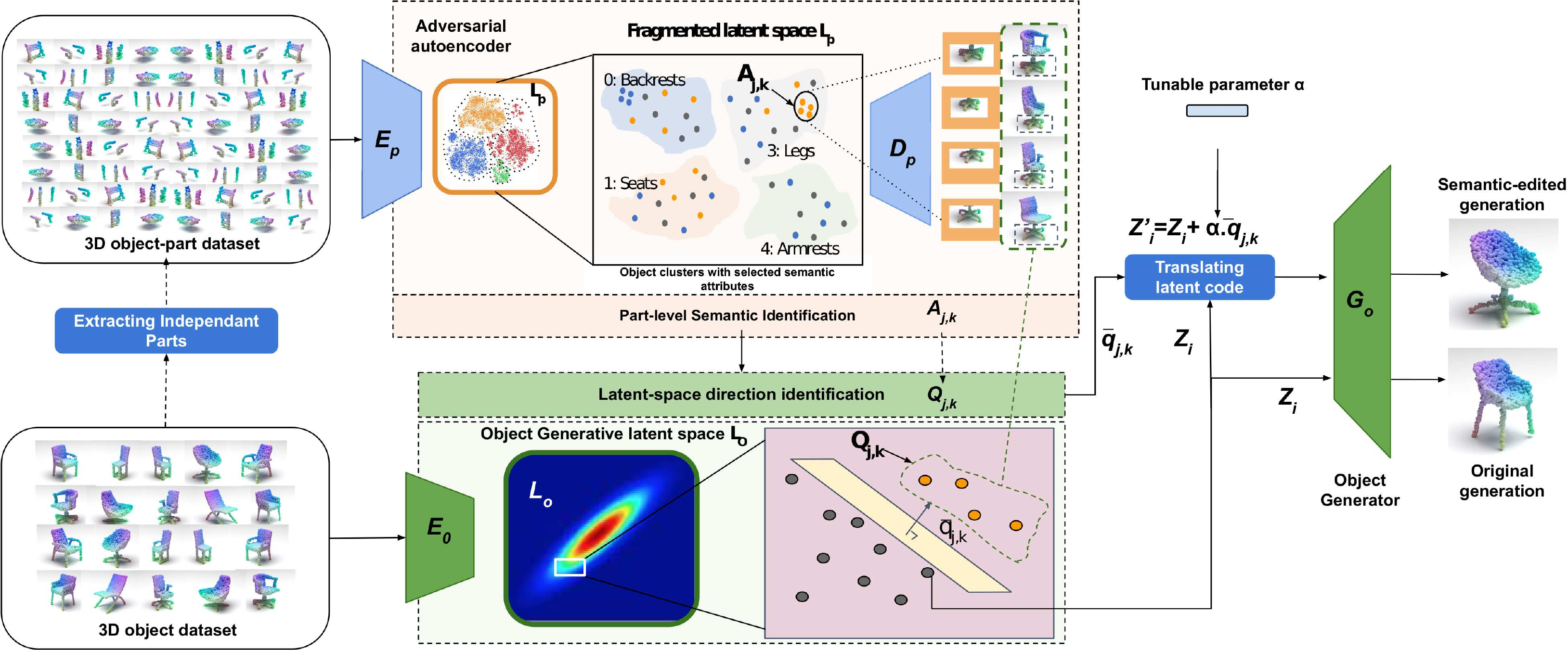}
  \caption{Architecture of the proposed framework. Fragmented latent space encodes the shapes of independent object parts and enable unsupervised part-level shape semantic identification. Identified semantics are combined with the contextual knowledge of object part connections from generative latent space to identify latent-space directions. Finally, latent codes of generator are edited by translating them in latent-space directions to generate objects with a required semantic attribute}
  \label{fig:architecture_diagram}
\end{figure*}
\section{The Framework}
\label{sec:framework}

In this section, we elaborate on our novel procedure to identify part-level shape-semantics using an object-part autoencoder latent space, and how we utilize those features to identify linear subspaces that encode part-level semantic features in generative latent spaces. 
Next, we illustrate how \(\name\) can translate generative latent codes towards a preferred semantic direction to enable part-level semantic-control in object generation. 
The architecture of the proposed framework is presented in Fig. \ref{fig:architecture_diagram}. 

\subsection{Identifying part-level semantic features and corresponding generative latent subspaces}

\paragraph{Part-level semantic features.} 
3D object datasets usually provide part annotations for objects \textit{(e.g. Backrest, seat etc.)}, and occasionally subclass labels within a superclass \textit{(e.g. Lounge chair, Armchair etc.)}. But there's a lack of annotated labels on different shape semantics of each part \textit{(e.g. Swivel legs, Reclined backrests etc.)}.
To mitigate this, we utilize a \textit{weakly supervised} approach to extract shape-semantics of object parts. 
(\(\name\) requires part-annotations for the object to train the part segmentation model, but it does not require any annotations for the shape semantic.)

Let $S_{i}$ be the $i^{th}$ point cloud of class $C$, consisting of $n$ different object parts. 
$S_{i} =\bigcup _{j=0}^{n} S_{ij}$ where $S_{ij}$ is the subset of points that represent the $j^{th}$ part category (Eg; Armrests of a chair). 
We leverage a 3D shape auto-encoder consisting of an Encoder $E_p$ and a Decoder $D_p$ to extract a latent space $L_p$. 
$z_{ij}=E_p(S_{ij}),  \bar{S_{ij}}= D_p(z_{ij})$ , $L_{p} =\bigcup _{i,j} z_{ij}$. 
The latent space $L_p$ contains information about
the shapes, sizes, and rotation information for different individual object parts. 
Although this autoencoder is not trained with part-labels, the resulting $L_p$ is fragmented with clear segregation between parts \textit{(e.g. Seats, Backrests, etc)}, indicating the structural knowledge acquired by the latent space. 
\emph{(We experiment with training separate latent spaces for separate parts, but the improvement was only marginal at the cost of added complexity and loss of scalability.)} 

We perform a hierarchical agglomerative clustering of part latent codes within $L_p$ separately for each object part $j$ to identify different part-level semantics. (The exact hyper-parameters of the clustering method is available along with the code implementation). A resulting cluster $A_{jk}$ contains the part-level latent codes of the $j^{th}$ part (Eg:Legs) that represent its $k^{th}$ unique semantic attribute (Eg:Swivel leg). (Note that all instances of one part category are treated as an ensemble; ie, all legs of the chair.)
The set $\{S_{i} |z_{ij} \in A_{jk}\}$ contains a set of point cloud objects that posses a specific semantic attribute $k$ localized in its $j^{th}$ part category \emph{(E.g- The set of chairs whose legs are swivel)}, irrespective of the geometries in the rest of the parts \emph{(Backrest, Seat, Armrest)} as shown in Fig. \ref{fig:architecture_diagram}. Since we have independently encoded the structural information of each part, these clusters show a diversity of features in the rest of the parts. 

\paragraph{Generative latent subspaces for part-semantics.} 
Point cloud object generators are trained to map a latent code $Z_i$ to a point cloud object $\widehat{S_{i}}= G_o(Z_i)$. These models are also often accompanied by a representation encoder $E_o$ to calculate the latent code $Z_i$ from a given point cloud object $S_{i}$, such that $Z_{i}= E_o(S_{i})$. Using this encoder, we form a set of latent codes within $L_o$ containing object-level embeddings for point clouds objects in our dataset \textit{(Shapenet Chair)}. Using the clusters $A_{jk}$ identified from the part-latent-space $L_p$, we find sets of positive examples (of object latent codes) from $L_o$ for each $k$th semantic attribute  localized in the $j$th object part  as $Q_{jk}=\{Z_{i} |z_{ij} \in A_{jk}\}$. Note that a latent code $Z_i$ could be a positive example for multiple semantic attributes, but in different object parts. Unlike for datasets like 2D face images \cite{liu2015celebA}
for which positive and negative semantic attributes can be easily defined \textit{(e.g. Smile-frown, Old-young etc.)}, assigning negative meanings for a shape semantic is difficult and ambiguous. Therefore, for every set of positive examples $Q_{jk}$, we define negative examples as $Q_{jk}'=\{Z_{i} |z_{ij} \notin A_{jk}\}$, that simply include all the latent codes $Z_i$ except those included in $Q_{jk}$.

\begin{table}[]
\scriptsize
\setlength{\tabcolsep}{4pt}
\renewcommand{\arraystretch}{1.2}
\caption{Average classification accuracy for linear SVMs fitted for each part-level semantic attribute in the object generative latent space. \emph{High accuracies confirm the existence of disentangled linear subspaces that encode part-level semantics.}}
\centering
\begin{tabular}{llllll}
\hline
\multirow{4}{*}{\textbf{Chair}}    &                       & \textbf{Backrest} & \textbf{Seat}   & \textbf{Legs}   & \textbf{Armrest} \\
                                   & \textbf{3DAAE}        & 0.957             & 0.975           & 0.967           & 0.917            \\
                                   & \textbf{Pointflow}    & 0.956             & 0.975           & 0.974           & 0.918            \\
                                   & \textbf{Diffusion P.} & 0.957             & 0.981           & 0.98            & 0.938            \\ \hline
\multirow{2}{*}{\textbf{Airplane}} & \textbf{}             & \textbf{Body}     & \textbf{Wings}  & \textbf{Tail}   & \textbf{Engine}  \\
                                   & \textbf{3DAAE}        & 0.88              & 0.854           & 0.868           & 0.854            \\ \hline
\multirow{2}{*}{\textbf{Car}}      & \textbf{}             & \textbf{Hood}     & \textbf{Bonnet} & \textbf{Wheels} & \textbf{Body}    \\
                                   & \textbf{3DAAE}        & 0.911             & 0.872           & 0.936           & 0.826            \\ \hline
\end{tabular}
\label{tab:SVM_results}
\end{table}

Next, we fit an independent linear SVM in the \emph{object generative latent space} $L_o$ for each identified semantic attribute using $Q_{jk}$ and $Q_{jk}'$ as positive and negative examples. We computed the classification accuracy for linear SVMs fitted for three generative latent spaces -- 3DAAE~\cite{Zamorski20203dAAE}, Pointflow~\cite{guandao2020pointflow}, and Diffusion P.~\cite{Shitong2021diffpm} -- leveraging ShapeNet~\cite{shapenet2015} dataset. Tab.\ref{tab:SVM_results} shows high classification accuracy irrespective of the generative model. Thus, we validate the hypothesis that \emph{there are linear subspaces in 3D generative latent spaces that embed disentangled representations for individual object part semantics}. While a similar observation had been made about the 2D binary facial semantics in \cite{Shen2020interfaceGAN}, this is the first analysis and experimental validation of the existences of non-binary part-level semantics in 3D generative latent spaces. \emph{(We also observed that creating clusters in the object generative space $L_o$ leads to heavy part-level semantic entanglements due to missing part knowledge. Hence, we utilize two different latent spaces; $L_p$ and $L_o$.)}

\begin{figure}[t]
  \centering
    \includegraphics[width=0.8\linewidth]{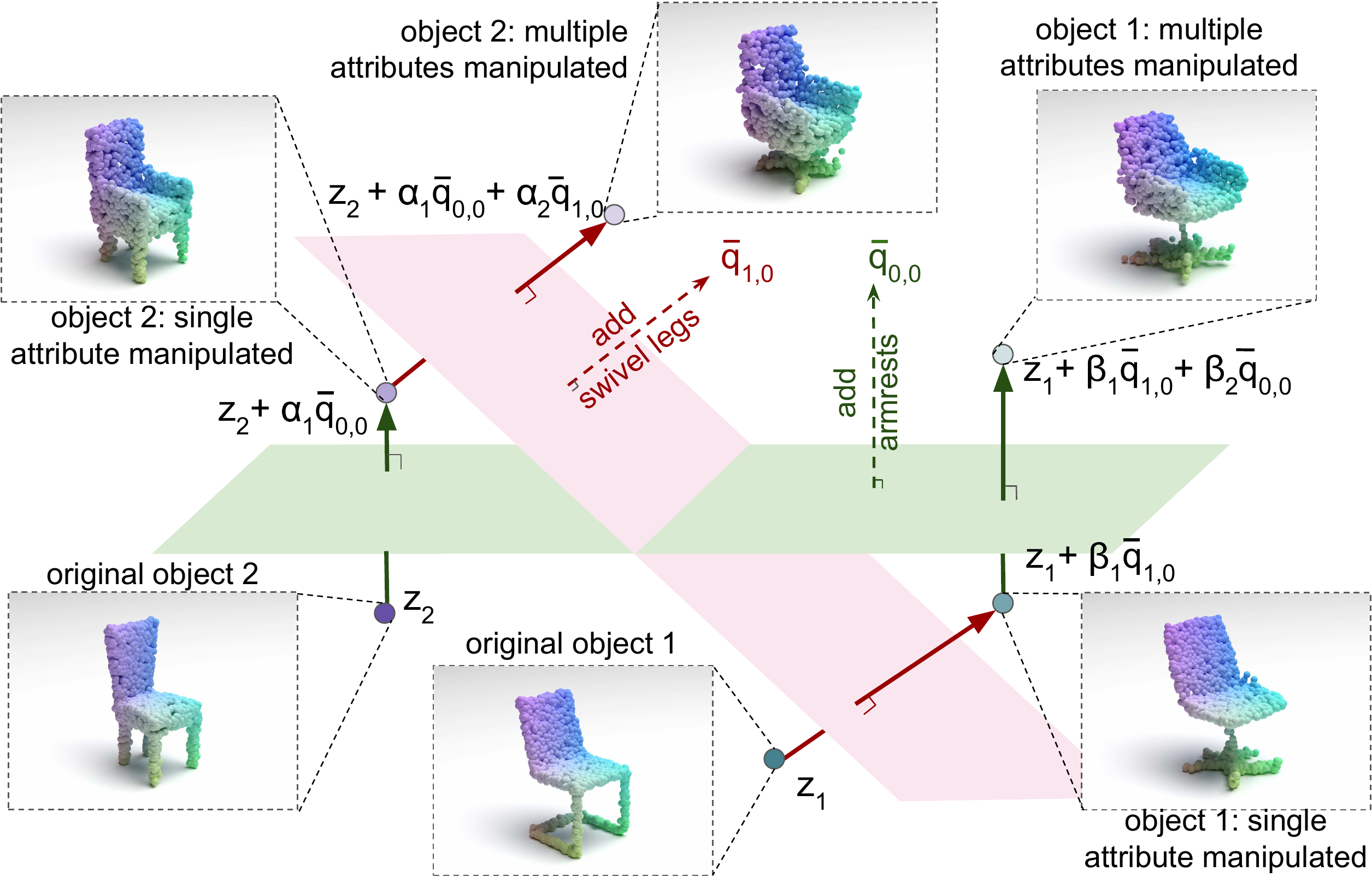}
  \caption{Given an object point cloud, our framework can manipulate either single or multiple semantics, specific to different object parts, through simple latent code translations. Latent codes are translated perpendicular to the hyperplanes fitted for the two part-semantics.}
  \label{fig:Teaser}
\end{figure}

\subsection{Semantic aware navigation in generative latent spaces}
The unit normal vector $\Bar{q}_{jk}$ to the hyperplane fitted by the SVM separating $Q_{jk}$ and $Q_{jk}'$ indicates a direction in the object generative latent space. We illustrate that we can manipulate the extent of the particular semantic ($k^{th}$ semantic localized in the $j^{th}$ object part) in the generated point cloud $\widehat{S_{i}}'$ by translating a latent code $Z$ in the direction of $\Bar{q_{jk}}$, $Z' = Z+\alpha \Bar{q}_{jk}$ where $\alpha$ is a tunable scalar parameter that can be used to control the level of semantic \emph{(E.g- the level of reclination of the chair's backrest)}. 
Moreover, this linear operation can be extended to combine multiple directions to simultaneously edit semantics in multiple object parts as shown in Fig.~\ref{fig:Teaser} and Eqn.~\ref{eqn1}. 
\begin{equation}
Z_{i} '=Z_{i} +\alpha _{0}\overline{q}_{o,k_{0}} +\alpha _{1}\overline{q}_{1,k_{1}} +...+\alpha _{n}\overline{q}_{n,k_{n}}
\label{eqn1}
\end{equation}

The positive example sets $Q_{jk}$ are not mutually exclusive for different object part categories $j$, hence, occasional entanglements across semantics in different object parts are possible. In the forthcoming sections, we show that such entanglements are caused by biases in the dataset, and sometimes are required to preserve the realistic nature of generated objects. 
Furthermore, the translated latent code $Z'$ is within $L_o$ itself, hence the same mapping function $G_o$ can be used for controlled object regeneration without any additional post-processing. 
The different part-level semantic features identified as clusters in the fragmented latent space corresponds to different features (like \textit{swivel legs, cantilever legs, straight legs etc.}) within a particular part, (\textit{leg}). 
The number of different semantic features identified for each part depends on the dataset. If the dataset has significant diversity in semantics, clustering will identify more unique semantic features, hence more latent-space directions and therefore, finer-grained semantic control could be enabled for object regeneration. However, we observe that \(\name\) can successfully identify latent directions with as few as 50 objects representing the particular part-semantic. (Clusters for part-semantics contain between 7\% to 50\% of all samples)

\section{Experiments}
In this section, we quantitatively and qualitatively evaluate the performance of \(\name\), and compare 
We introduce two novel  metrics to quantify \emph{consistency} and \emph{ability to localize} semantic manipulations to object parts
and show that \(\name\) outperforms the existing disentanglement methods.

\subsection{Evaluation setup}
\label{sec:eval_setup}

\noindent
\textbf{Dataset.} We use 3D models of ShapeNet~\cite{shapenet2015} dataset to obtain a pointcloud representation of $2048\times3$ for each object and use them to train the object generative models. Moreover, we utilize the point-level part annotations provided by ShapeNet-part\cite{Yi2016shapenet_part} dataset to train a PointNet\cite{charles2017pointnet} based object segmentation model to create part decompositions required for the \(\name\) pipeline.

\noindent
\textbf{Metrics.}  

\textbf{(i). Semantic Localization Score (SLS).} To ensure controllability of part-level object manipulation, it is required that the change added to the object is restricted to the intended part. We introduce the ratio of structural change added to the intended part and to the rest of the object parts as a novel metric, \textit{Semantic Localization Score (refer Eqn. \ref{eq:SLS}).} \emph{A higher value of SLS indicates better controllability of part level semantic manipulations.}
\begin{equation}
\small
\scriptstyle
\label{eq:SLS}
    SLS_{k,a} =E\left[\frac{CD\left(\widehat{S_{ia}} ,\widehat{S_{ia}} '\right)}{CD\left(\bigcup _{j\neq a}\widehat{S_{ij}} ,\bigcup _{j\neq a}\widehat{S_{ij}} '\right)}\right]
\normalsize
\end{equation}

$CD$ is the Chamfer distance \cite{DPDist20} between original pointcloud $\widehat{S_{i}}$ and manipulated pointcloud $\widehat{S_{i}}'$ by translating latent code $Z_i$ towards a particular latent-space direction; $Z' = Z+\alpha \Bar{q}$.
$\widehat{S_{ij}}$ is the set of points that represent the $j^{th}$ part of the object. 
Expected value $E$ is calculated by translating 1000 random latent samples by an empirically selected distance $\alpha$ towards a particular latent direction.

\textbf{(ii). Semantic Consistency Score (SCS).} Consistency of part-level object manipulation is achieved if all different latent codes translated towards a particular direction is consistently added with the same semantic change.
We quantify the consistency by using the confidence score of a semantic classifier model for the transformed object. 
For a given part semantic (e.g., swivel leg), we manually label the objects with the given part semantic as positive labels and the objects with other semantic attributes (e.g., straight leg, cantilever leg) as negative labels. Then we train a set of binary semantic classifier models (Eg: Straight leg or not a straight leg) with a PointNet \cite{charles2017pointnet} feature extractor. 
The model is expected to classify the transformed object to the corresponding semantic category. For example, a model trained using \textit{swivel legs} as positive labels should classify a latent navigation towards \textit{swivel legs} direction as positive outputs and that of \textit{cantilever legs} and \textit{straight legs} directions as negative outputs. 
We use a well-validated PointNet model to perform this evaluation which yields the consistency of semantic manipulations quantitatively.

\noindent
\textbf{Comparison.} We compare the performance of \(\name\) in identifying latent directions with other unsupervised latent disentanglement methods. \textit{Closed-form} 
\cite{Shen2021ClosedFormFO} proposes a closed-form factorization based method to find semantically meaningful directions in a GAN latent spaces in a fully unsupervised manner by directly decomposing the pre-trained weights of the generator. \textit{GANSpace}~\cite{GANSpaceHarkonenHLP20} proposes a technique to analyze GANs and create interpretable controls in terms of latent directions based on a Principal Component Analysis (PCA). 
Both have only been applied to 2D image synthesis tasks by the original authors. 
We adapt these approaches to 3D object synthesis using a 3DAAE\cite{Zamorski20203dAAE} model and compare its performance with \(\name\) in terms of identifying disentangled latent directions for part semantic control. 
Both approaches yield a large number of directions that mostly added either the same semantic change or no change at all.
To compare \(\name\) with their upperbound performances, we matched the
latent directions from these methods (out of 100+ suggested directions) with those of \(\name\) after manual observation. 

The part-editing performance of \(\name\) is limited by the representation capability of the selected generative model. 
Hence, we cannot fairly compare the part-editing performance with specialized models such as \cite{anastasia2019composite, kaichun2020Structedit}. We focus on the performance of latent disentanglement of \(\name\) where our key contribution is. 

\subsection{Results}
\label{sec:results}

\begin{figure}[t]
  \centering
  
  \includegraphics[width=1\linewidth]{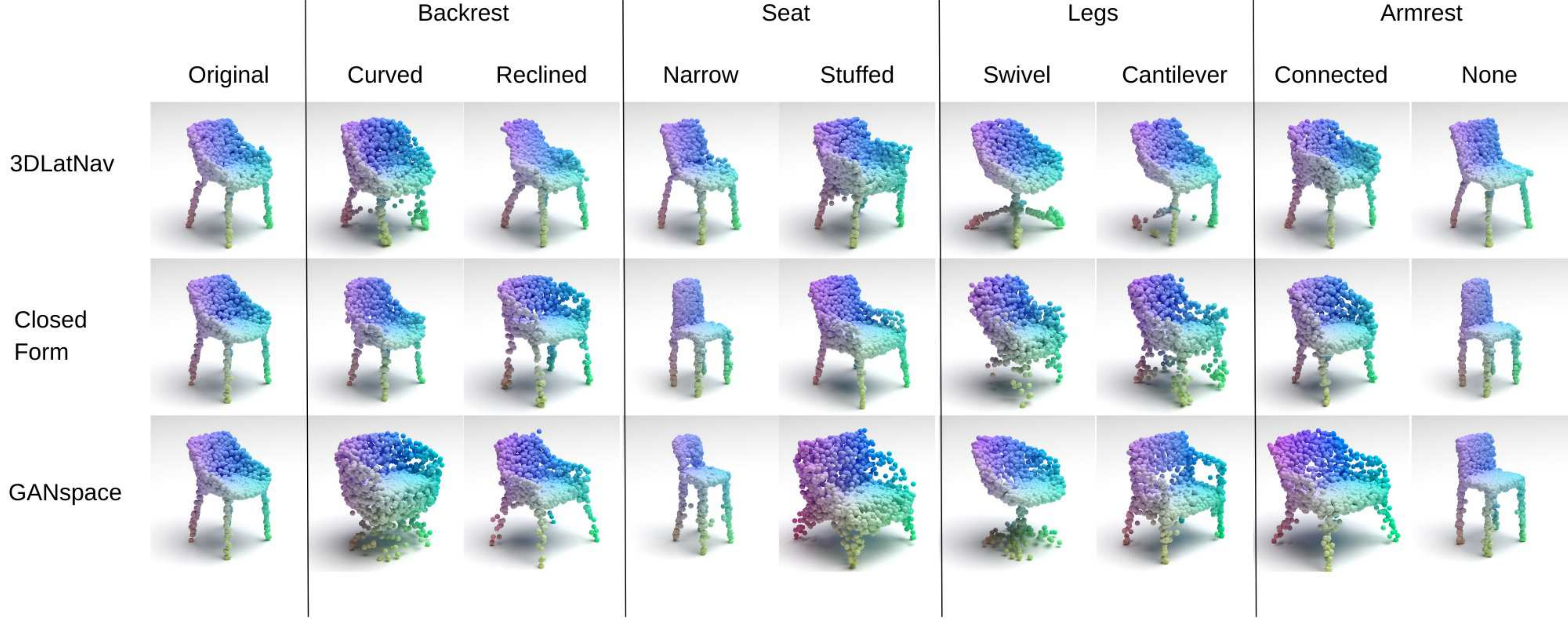}
  \caption{Comparison of part-level object manipulation results of \(\name\) with unsupervised latent disentanglement methods; Closed-form\cite{Shen2021ClosedFormFO} and GANSpace\cite{GANSpaceHarkonenHLP20}. 
}
  \label{fig:comparison_visual}
\end{figure}

\noindent
\textbf{Visual qualitative results.} Fig.~\ref{fig:single_attrib} illustrates the quality of semantic manipulations of \(\name\) for different object parts and semantic features. Fig.\ref{fig:comparison_visual} shows results of part level semantic manipulation using \(\name\), GANSpace~\cite{GANSpaceHarkonenHLP20} and Closed-form~\cite{Shen2021ClosedFormFO}. 
Unlike in \(\name\), for a majority of the cases both of the other methods add unintended changes to the other parts of the objects during object manipulation. (E.g- \emph{Curved backrest and Stuffed seat in GANSpace} adds changes legs, \emph{Connected armrest in GANSpace} adds changes to seat, \emph{Removing armrest in Closed-form} adds changes to backrest, etc.) 
Although \emph{GANSpace} has been more successful than \emph{Closed-form} in identifying directions for common semantics like \emph{swivel legs}, specific semantics like \emph{cantilever legs} were not present/clearly distinguishable among directions yielded by either GANSpace or Closed-form. 
\(\name\) can be used on the generative latent spaces of any pretrained generator. 
\textit{We present the visual results of applying our method on different generators \cite{Zamorski20203dAAE, guandao2020pointflow, Shitong2021diffpm} in supplementary materials.}

\begin{table}[h]
\scriptsize
\centering
\setlength{\tabcolsep}{1.1pt}
\renewcommand{\arraystretch}{1.2}
\caption{Semantic localization score (Eqn. \ref{eq:SLS}) to quantify the relative structural change added to the specific object part by manipulating a semantic attribute.}
\begin{tabular}{l|c|c|ccccc}
\hline
\textbf{\begin{tabular}[c]{@{}l@{}}Disentanglem\\ -ent method\end{tabular}} & \multicolumn{1}{l|}{\textbf{\begin{tabular}[c]{@{}l@{}}Generative \\ model\end{tabular}}} & \multicolumn{1}{l|}{\textbf{\begin{tabular}[c]{@{}l@{}}Object \\ class\end{tabular}}} & \multicolumn{1}{l}{\textbf{Part1}} & \multicolumn{1}{l}{\textbf{Part2}} & \multicolumn{1}{l}{\textbf{Part3}} & \multicolumn{1}{l}{\textbf{Part4}} & \multicolumn{1}{l}{\textbf{Average}} \\ \hline
\multirow{4}{*}{\(\name\)}                                                   & \multirow{4}{*}{\textbf{3DAAE}}                                                           & \multirow{2}{*}{airplane}                                                             & \textbf{Body}                      & \textbf{Wings}                     & \textbf{Tail}                      & \textbf{Engine}                    &                                      \\
                                                                            &                                                                                           &                                                                                       & 0.61                               & 0.7513                             & 2.5914                             & 5.3587                             & 5.4342                               \\ \cline{3-8} 
                                                                            &                                                                                           & \multirow{2}{*}{car}                                                                  & \textbf{Hood}                      & \textbf{Bonnet}                    & \textbf{Wheels}                    & \textbf{Body}                      & \textbf{}                            \\
                                                                            &                                                                                           &                                                                                       & 6.372                              & 3.3651                             & 2.6592                             & 0.4635                             & 3.6267                               \\ \hline
\multirow{4}{*}{\(\name\)}                                                   & \textbf{}                                                                                 &                                                                                       & \textbf{Backrest}                  & \textbf{Seat}                      & \textbf{Legs}                      & \textbf{Armrest}                   &                                      \\
                                                                            & \textbf{Pointflow}                                                                        & chair                                                                                 & 3.885                              & 0.329                              & 7.129                              & 10.113                             & 5.403                                \\ \cline{2-8} 
                                                                            & \textbf{Diffusion}                                                                        & chair                                                                                 & 1.826                              & 1.121                              & 6.425                              & 7.774                              & 4.263                                \\ \cline{2-8} 
                                                                            & \textbf{3DAAE}                                                                            & chair                                                                                 & \textbf{1.6943}                             & 0.4703                             & \textbf{8.7261}                             & \textbf{7.8573}                             & \textbf{4.7617}                               \\ \hline
GANSpace                                                                   & \textbf{3DAAE}                                                                            & chair                                                                                 & 1.5274                             & \textbf{0.595}                              & 6.3166                             & 3.1231                             & 2.8905                               \\ \hline
Closed form                                                                 & \textbf{3DAAE}                                                                            & chair                                                                                 & 1.403                              & 0.474                              & 4.453                              & 6.9471                             & 3.3192                               \\ \hline
\end{tabular}
\label{tab:SLS_score}

\end{table}

\noindent
\textbf{Semantic localization.} 
The average Semantic Localization Scores (SLS) greater than one for of the object parts in Tab.\ref{tab:SLS_score} show that the structural changes added to the intended part significantly exceeds those added to the rest of the object. There are also certain cases where the SLS is less than 1, but they account to the scenarios where a structural change to one part affects the rest of the parts, \textit{E.g- changing the seat of a chair requires a change in backrests, armrests, and legs.}
As shown in Table \ref{tab:SLS_score}, \(\name\) outperforms both GANSpace and Closed-form
in restricting the semantic changes to a required part, confirming its superior performance in disentangling part-level semantic controls. These results are agreeing with the visual results in Fig.\ref{fig:comparison_visual}. 
We also present results for implementing \(\name\) on Pointflow\cite{guandao2020pointflow} and Diffusion Probablistic models\cite{Shitong2021diffpm} generative latent spaces and confirm that the discovered latent directions can successfully control the localization of part-level object manipulations with different generative models.

\begin{table}[h]
\tiny
\centering
\setlength{\tabcolsep}{1.8 pt}
\renewcommand{\arraystretch}{1.2}

\caption{SCS to evaluate consistency of semantic manipulation. Transformed objects are tested with the trained binary classifier model. \(\name\) consistently outperforms the existing approaches in both object-level and part-level.}

\begin{tabular}{@{}l|cccccccccccc@{}}
\toprule
                                                      & \multicolumn{12}{c}{Object-Level}                                                                                                                                                                                                                                                                                                                                                                                                                            \\ \midrule
                                                      & \multicolumn{3}{c|}{Backrest}                                                                                 & \multicolumn{3}{c|}{Seat}                                                                                     & \multicolumn{3}{c|}{Legs}                                                                                     & \multicolumn{3}{c}{Armrest}                                                                                  \\ \midrule
                                                      & \rotatebox{90}{curved}                             & \rotatebox{90}{reclined}                           & \multicolumn{1}{c|}{\rotatebox{90}{stuffed}}        & \rotatebox{90}{narrow}                             & \rotatebox{90}{stuffed}                            & \multicolumn{1}{c|}{\rotatebox{90}{wide}}           & \rotatebox{90}{swivel}                             & \rotatebox{90}{straight}                           & \multicolumn{1}{c|}{\rotatebox{90}{cantilever}}     & \rotatebox{90}{connected}                          & \rotatebox{90}{disconnected}                       & \rotatebox{90}{none}                               \\
\begin{tabular}[c]{@{}l@{}}GANSpace\end{tabular}   & 0.705                              & 0.939                              & \multicolumn{1}{c|}{0.793}          & 0.620                              & 0.612                              & \multicolumn{1}{c|}{0.655}          & 0.933                              & 0.661                              & \multicolumn{1}{c|}{0.590}          & 0.604                              & 0.654                              & 0.919                              \\
\begin{tabular}[c]{@{}l@{}}Closedform\end{tabular} & 0.493                              & 0.683                              & \multicolumn{1}{c|}{0.468}          & 0.414                              & 0.665                              & \multicolumn{1}{c|}{0.663}          & 0.671                              & 0.573                              & \multicolumn{1}{c|}{0.684}          & 0.698                              & 0.656                              & 0.775                              \\
\begin{tabular}[c]{@{}l@{}}3DLatNav\end{tabular}   & \multicolumn{1}{l}{\textbf{0.943}} & \multicolumn{1}{l}{\textbf{0.980}} & \multicolumn{1}{l|}{\textbf{0.923}} & \multicolumn{1}{l}{\textbf{0.661}} & \multicolumn{1}{l}{\textbf{0.854}} & \multicolumn{1}{l|}{\textbf{0.764}} & \multicolumn{1}{l}{\textbf{0.851}} & \multicolumn{1}{l}{\textbf{0.888}} & \multicolumn{1}{l|}{\textbf{0.954}} & \multicolumn{1}{l}{\textbf{0.927}} & \multicolumn{1}{l}{\textbf{0.952}} & \multicolumn{1}{l}{\textbf{0.989}} \\ \midrule
                                                      & \multicolumn{12}{c}{Part-Level}                                                                                                                                                                                                                                                                                                                                                                                                                              \\ \midrule
\begin{tabular}[c]{@{}l@{}}GANSpace\end{tabular}   & 0.756                              & 0.906                              & \multicolumn{1}{c|}{0.734}          & 0.452                              & 0.702                              & \multicolumn{1}{c|}{0.673}          & \textbf{0.908}                     & 0.682                              & \multicolumn{1}{c|}{0.643}          & 0.410                              & 0.500                              & 0.820                              \\
\begin{tabular}[c]{@{}l@{}}Closed form\end{tabular} & 0.580                              & 0.677                              & \multicolumn{1}{c|}{0.409}          & 0.380                              & 0.664                              & \multicolumn{1}{c|}{0.658}          & 0.679                              & 0.540                              & \multicolumn{1}{c|}{0.674}          & 0.540                              & 0.440                              & 0.853                              \\
\begin{tabular}[c]{@{}l@{}}3DLatNav\end{tabular}   & \textbf{0.948}                     & \textbf{0.989}                     & \multicolumn{1}{c|}{\textbf{0.960}} & \textbf{0.692}                     & \textbf{0.869}                     & \multicolumn{1}{c|}{\textbf{0.719}} & 0.834                              & \textbf{0.737}                     & \multicolumn{1}{c|}{\textbf{0.923}} & \textbf{0.896}                     & \textbf{0.912}                     & \textbf{0.998}                     \\ \bottomrule
\end{tabular}
\label{tab:SCS_score_part}
\end{table}

\noindent \textbf{Semantic consistency.} 
The results of semantic consistency score (SCS) are reported in Tab. \ref{tab:SCS_score_part}. 
We compute object-level scores (Tab. \ref{tab:SCS_score_part} top) where the whole object is assessed with the trained model and the part-level scores (Tab. \ref{tab:SCS_score_part} bottom) where only the specific object parts are assessed with a trained part-level semantic classifier model. \(\name\) consistently outperforms the previous other latent disentanglement and navigation approaches\cite{GANSpaceHarkonenHLP20, Shen2021ClosedFormFO} in most semantic manipulations. 
These are supported by visual results in Fig. \ref{fig:comparison_visual} where \(\name\) shows consistency in the semantic manipulation across all semantics compared to the prior approaches and the visual results in Fig. \ref{fig:single_attrib} where \(\name\) shows consistency in semantic manipulations for different objects for a given latent direction.


\section{Analysis and discussion}
\label{sec:experiments}

We present an in-depth analysis of the properties associated with part-semantic representations in generative latent spaces and their manipulations using the discovered latent directions from \(\name\). Unless mentioned otherwise, \(\name\) is implemented with two 3D Adversarial Autoencoders (3DAAE), one as the object part shape autoencoder, and other as the 3D object generator.
\footnote{\label{note1}We will present specific details of implementation, experimental setups, and additional visual results in Supplementary materials.}

\subsection{Semantic entanglement of features}
\begin{figure}[h]
  \centering
  \includegraphics[width=0.8\linewidth]{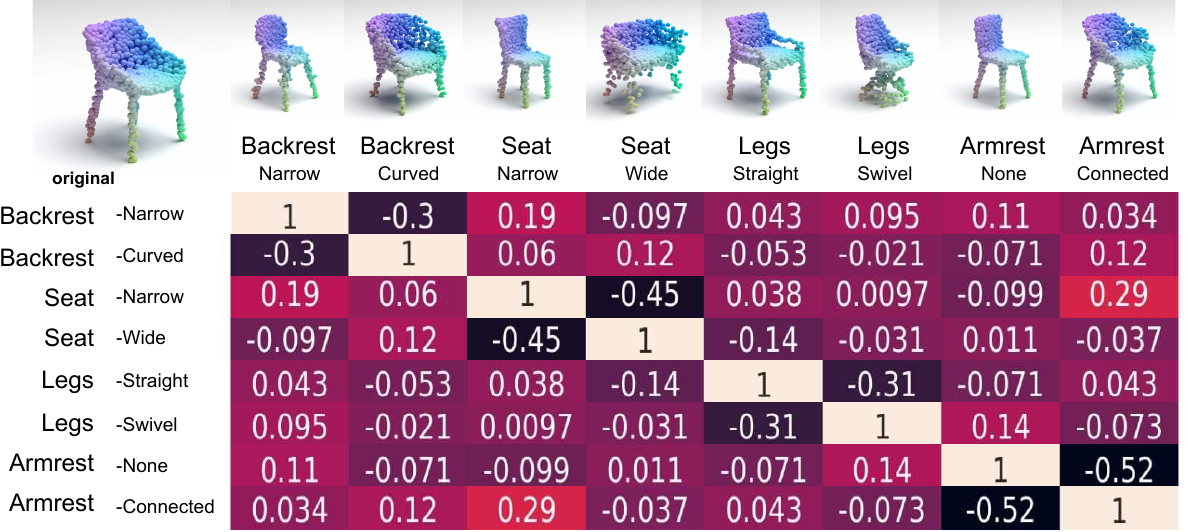}
  \caption{Cosine similarities between latent space directions each pair of selected part-semantics for Chair class.}
  \label{fig:cosine_similarity}
\end{figure}

We calculate Cosine similarity between identified $n$-dimensional latent space directions that encode different part-semantic attributes. Fig.\ref{fig:cosine_similarity} illustrates that entanglement of features across different object parts are very low with overall low Cosine similarity. Occasional entanglements such as between \textit{Narrow seats} and \textit{Armrests absent} preserve the realistic nature of the objects enforced by the ShapeNet\cite{shapenet2015} dataset.
Negative correlations are sometimes observed between semantically contrasting attributes of the same part, \emph{(E.g. highest negative correlation is between availability and absence of armrest.)}

\begin{figure}[h]
  \centering
    \includegraphics[width=0.7\linewidth]{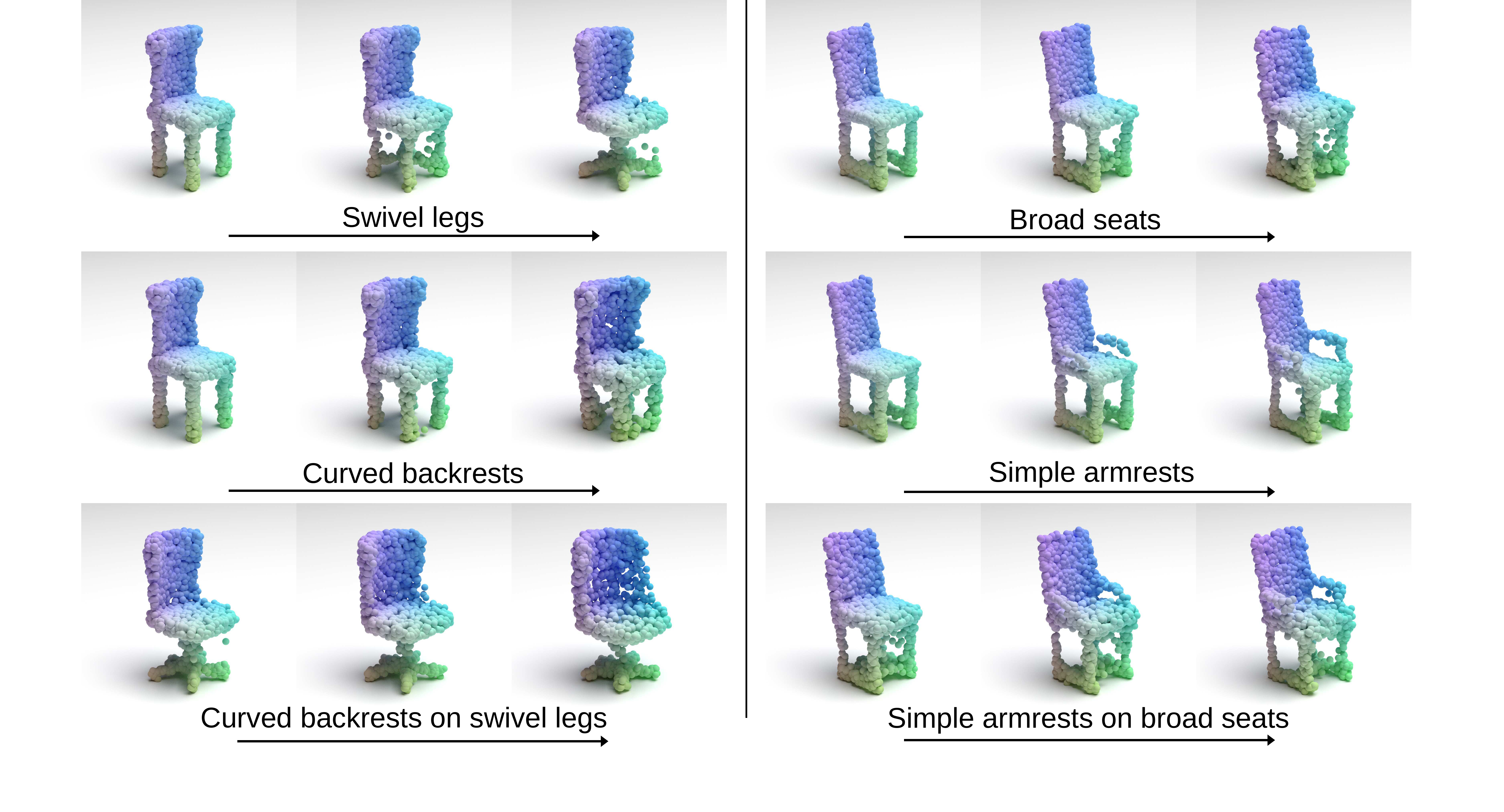}
  \caption{Manipulating multiple semantics by translating latent codes to a combination of latent-space directions.}
  \label{fig:multiple_semantic}
\end{figure}

\subsection{Part manipulation by latent space navigation}

\noindent
\textbf{Manipulating multiple semantics.} 
The linearity of the subspaces that encode part-level semantics enable us to manipulate/edit a query object with any linear combinations of identified semantics as formulated in Eqn.~\ref{eqn1}. For example, we can translate a latent code $Z$ in multiple directions $\Bar{q}_{0,1}$. and $\Bar{q}_{1,1}$ with a controllable extent of the added semantic using the tunable parameters $\alpha_1$ and $\alpha_2$, such that $Z' = Z+\alpha_1 \Bar{q}_{0,1} +\alpha_2 \Bar{q}_{1,1}$. The first two rows of Fig.\ref{fig:multiple_semantic} show individual semantic manipulation and the bottom row shows the linear combination of the two semantics while preserving the other features to a large extent.

\begin{figure}[t]
  \centering
    \includegraphics[width=1\linewidth]{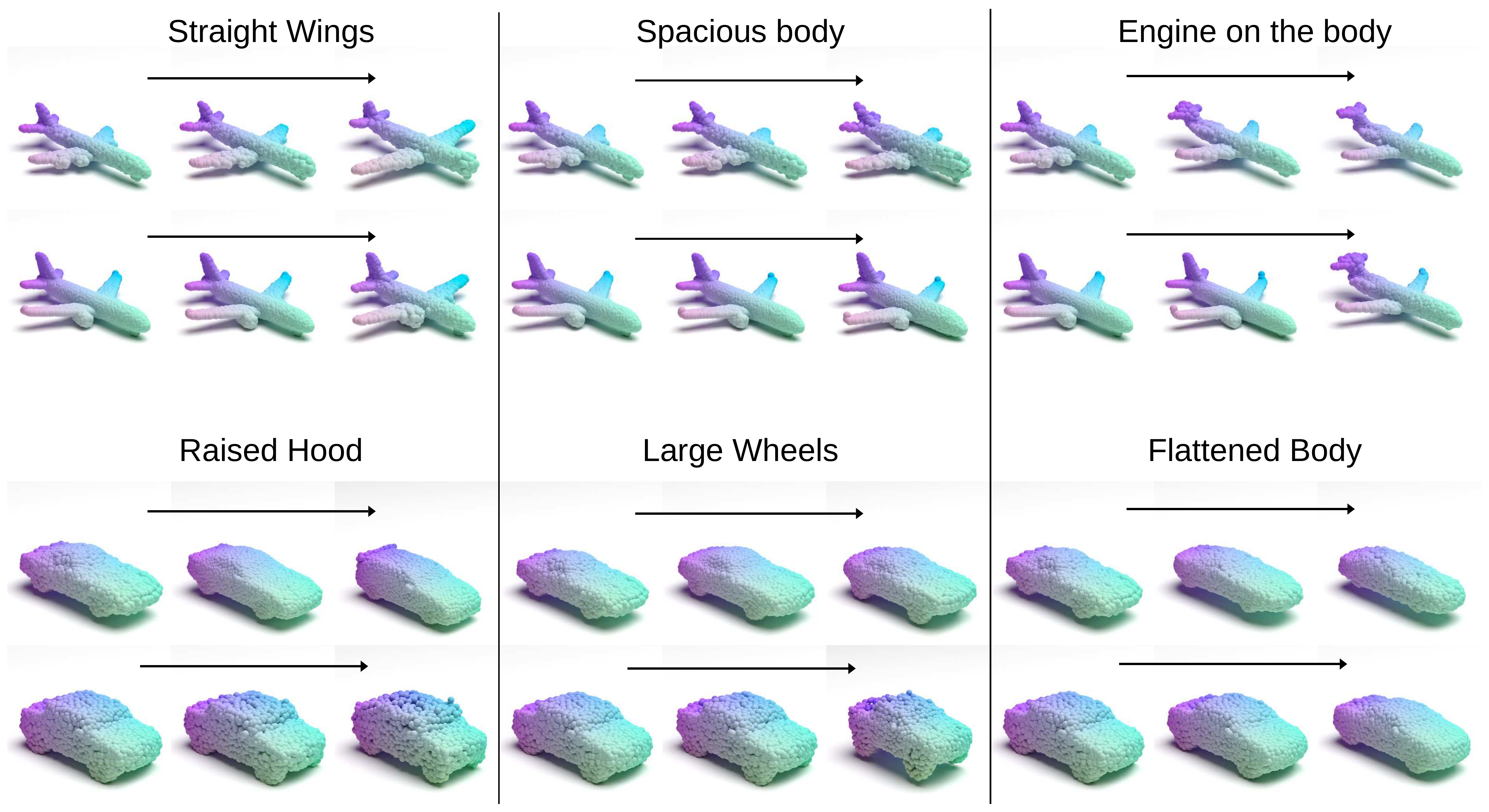}
  \caption{\(\name\) on other object classes. 
  Semantics can be controlled for different parts independently. 
  Feature entanglements are present where realistic nature of the object needs to be preserved.
  \textit{e.g., changing the airplane tail when engine is moved from wings to the body.
  }  }
  \label{fig:cars_aeroplanes}
\end{figure}

\noindent
\textbf{Quality of generated and transformed objects.} 
We demonstrate the robustness of \(\name\) by conducting semantic manipulation for different object classes in Fig.~\ref{fig:cars_aeroplanes}
The results show that \(\name\) semantic manipulation performs similar to the chair class in other two classes of airplane and car. Unlike 3D object part manipulation/editing frameworks that detach the object parts to interchange or interpolate between different objects and subsequently combine them to form the complete object, \(\name\) generates semantically edited objects from the same generator with acquired contextual knowledge of object part connections \cite{anastasia2019composite,cihan2021LPMnet}. 
Therefore, \(\name\) does not produce disconnected object parts or unrealistic object shapes. 
The edited point clouds are guaranteed to have the quality offered by the generative model. 

\subsection{Further analysis}
A generator trained on a certain dataset may  have entangled part-semantic features due to correlations in the dataset. To ablate this effect, we trained a generator on an extended ShapeNet dataset including additional sets of examples synthesized by naively combining objects parts from different objects. We show that the directions found using \(\name\) are even more disentangled when the generative model has been trained with such synthetic data as opposed to limiting to real-life objects. We present these results and the implementation details in supplementary materials.


\section{Conclusion and future work}
We introduced \(\name\), a framework to identify meaningful directions in 3D generative latent spaces that independently encode semantics of different object parts.  Using comprehensive qualitative and quantitative measurements, we show that \(\name\) outperform existing unsupervised latent disentanglement methods in identifying meaningful latent directions.
\(\name\) can successfully be used to manipulate part level semantics of 3D objects using any pretrained generative model, without any post processing or reference objects to extract the target features from. \emph{We contribute to widening of understanding of how 3D part semantics behave in generative latent spaces by presenting extensive analysis and ablation studies.}
Further research directions will be measuring the utility of \(\name\) as an editing framework (using a user-study) to create a 3D semantic manipulation framework deployable practically in interactive mixed reality or gaming applications, and extending the framework to other types of 3D data representations
and to larger 3D scenes.

%
%

\section{Code and Demo}
We have made the \href{https://github.com/theamaya/3DLatNav}{project repository} and an \href{https://colab.research.google.com/drive/1w1QoyfIkhWFQG1wvGgpGRhVivn4-LNu9?usp=sharing
}{interactive demo} publicly available. 
\footnote{Repository- \href{https://github.com/theamaya/3DLatNav}{https://github.com/theamaya/3DLatNav},

\noindent Demo- \href{https://colab.research.google.com/drive/1w1QoyfIkhWFQG1wvGgpGRhVivn4-LNu9?usp=sharing}{https://colab.research.google.com/drive/1w1QoyfIkhWFQG1wvGgpGRhVivn4-LNu9?usp=sharing}}

\section{Implementation Details}

\subsection{\(\name\) framework}

We used the ShapeNet~\cite{shapenet2015} dataset throughout this work. First to extract part-level semantic features, we part-segmented the point clouds of Shapenet dataset using a PointNet~\cite{charles2017pointnet} part-segmentation model trained on ShapeNet-Part dataset~\cite{Yi2016shapenet_part}.
To obtain the object-part latent space $L_p$ for these segmented parts, we used the 3D Adversarial Autoencoder architecture~\cite{Zamor20203dAAE}. We train a 3DAAE model with different parts of a particular object class using input point clouds of 400 points to generate a 400-dimensional latent space $L_p$. Each individual object part was either up-sampled or down-sampled to yield a representation of 400 points. The unsupervised semantic feature identification explained in Sec. 3 of the main paper is done on this latent space.

We verified \(\name\) for multiple generative models and their corresponding generative latent spaces. Baseline results were generated for a 3DAAE model trained for a specific object class on the original ShapeNet dataset. The input point clouds are generated by sampling 2048 points from each object. The model is trained to regenerate 2048 points using a 2048-dimensional latent space $L_o$. Correspondingly other generative models~\cite{guandao2020pointflow, Shitong2021diffpm} were also trained on the same train dataset. The object generative latent space $L_o$ is generated by extracting the encodings of each object from the Shapenet dataset using their corresponding representation encoders. The dimensions of each latent space varies as per defaults set by original authors. 

Fitting SVMs to find linear subspaces, deriving of latent directions and point cloud part-editing by translating latent codes as mentioned in Sec.3 of the main paper is carried out on this latent space $L_o$.

\subsection{Quantitative evaluation}
In this work we introduced two quantitative metrics to evaluate the performance of part-level semantic manipulation by navigating object generative latent spaces. The two metrics quantify two properties of the resulting manipulated regenerations.

\noindent \textbf{Semantic localization score (SLS).} 
As shown in Eqn.2 of the original submission, we use Semantic Localization Score (SLS) to measure the level of localization of the intended manipulation to the specific object part as compared to the rest of the object.
To calculate SLS for a single object manipulation, we part-segment both the original object and the manipulated object, and create two point sets each; points representing the object part that we plan to manipulate, and the points representing the rest of the object. Next we calculate the Chamfer distance\cite{DPDist20} between the corresponding pointsets from the original and manipulated objects, and calculate their ratio to obtain the SLS. The average results reported in Tab.2 of the original submission is calculated by translating 1000 random objects to each meaningful direction discovered by the particular framework, and averaging the results over all objects and all directions corresponding to each object part. 
The $\alpha$ (distance translated in the latent space $L_o$) is selected empirically for each generative latent space.

\noindent \textbf{Semantic consistency score (SCS).} 
As reported in Tab.3 of our main paper, we calculated semantic consistency scores to evaluate the consistency of \(\name\)  transformations on different objects. We train 12 binary semantic classifier models for object-level and part-level evaluations each. (3 most prominent semantics for each part- \textit{Legs- swivel, cantilever, straight, Armrests- connected, disconnected, none, Backrest- narrow, stuffed, reclined, Seats- narrow, wide, stuffed}). The binary classifiers were trained using manually annotated objects from the shapenet dataset with the above part-level semantics. (\textit{Eg- For a swivel leg binary classifier, shapenet chairs with swivel legs were taken as positive examples while those with straight and cantilever legs were taken as negative examples.})
We transformed a set of 1000 randomly selected chairs to each of these 12 directions to create the test set. The average results for these 1000 objects are reported in Tab.3. Similar to the evaluation of Semantic Localization Score, the magnitude of transformation $\alpha$ was selected empirically for different generative latent spaces.

\section{Further Results}
\begin{figure}[h]
  \centering
    \includegraphics[width=1\linewidth]{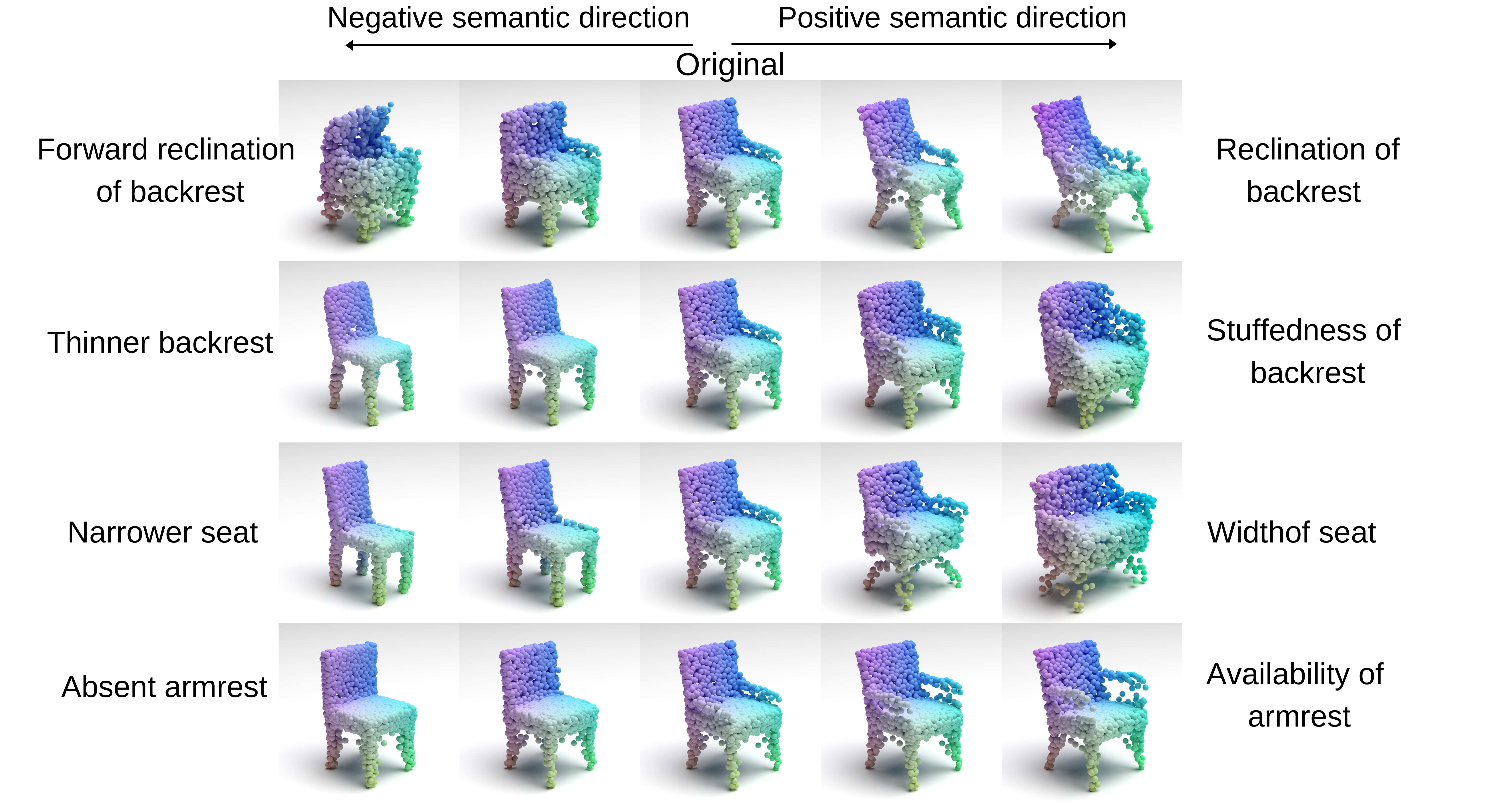}
  \caption{Emergence of negative attributes when translating latent codes to the negative directions.}
  \label{fig:negative_attrib}
\end{figure}

\textbf{Emergence of negative attributes.} 
The latent space directions are calculated based on hyperplanes that separate a set of positive examples $Q_{jk}$ of a selected part-level shape semantic against the rest of the training dataset, $Q_{jk}'$. It should be noted that the set $Q_{jk}'$ does not exclusively contain examples with an opposite semantic attribute, but includes all the objects in which the considered semantic attribute is not present.
Nonetheless, upon translating a latent code to the negative direction of an identified latent-space direction, a corresponding interpretable negative semantic attribute is added to the generated objects as shown in Fig.\ref{fig:negative_attrib}. 
It shows that latent spaces of generative models have embedded information about matching negative semantics for a specified part-level semantics, \emph{although specific negative examples were not provided during training the SVMs}.

\begin{figure*}
  \includegraphics[width=1\linewidth]{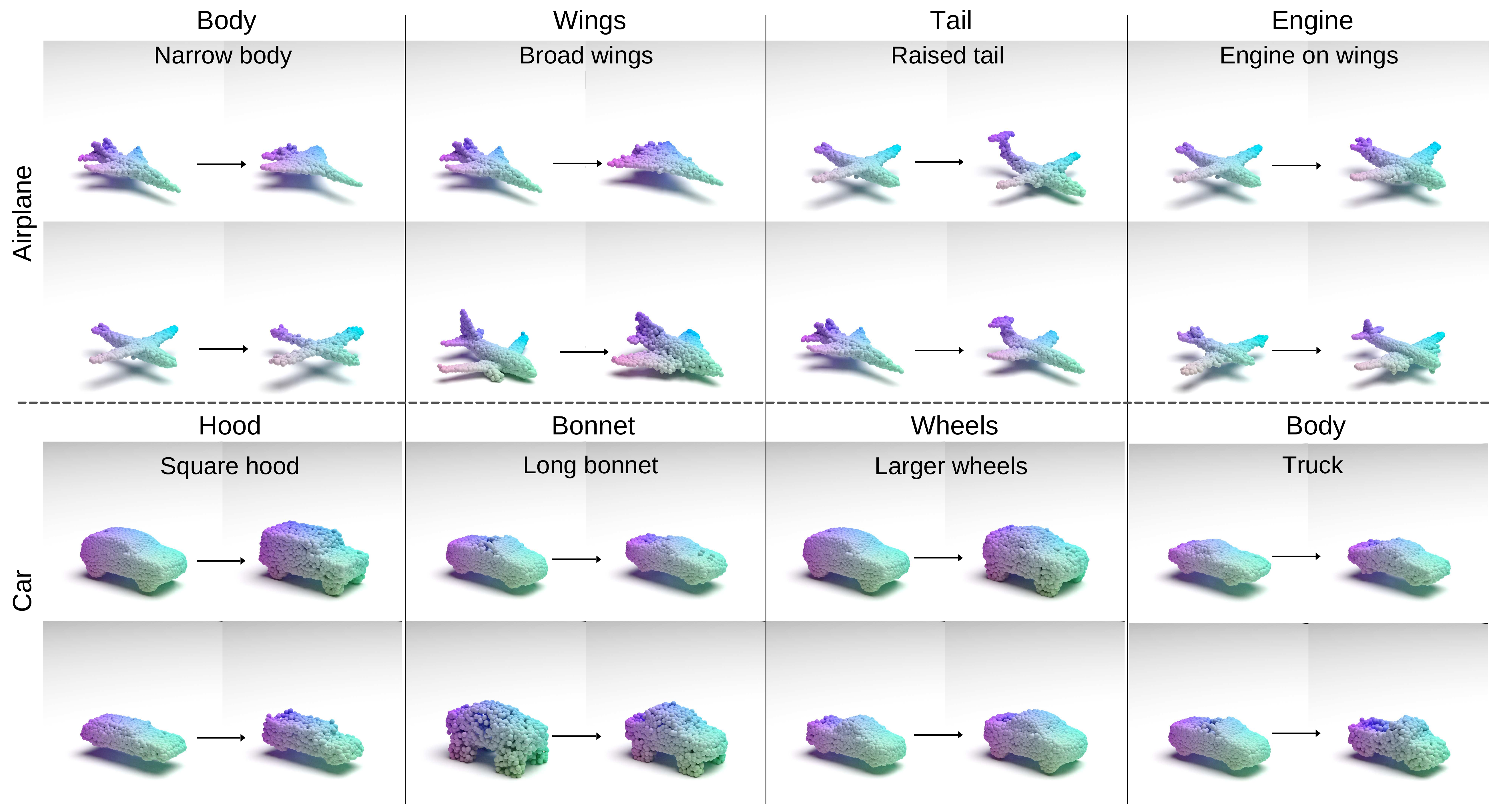}
  \caption{Manipulating single semantic attributes by translating a latent code to the identiﬁed latent space directions in Airplane and Car classes. The part-level features change while preserving the other parts.}
  \label{fig:airplane_car_single_sem}
\end{figure*}

\noindent
\textbf{Model robustness on other object classes}
The proposed framework can successfully identify and manipulate part-level semantics in different object classes. Fig.\ref{fig:airplane_car_single_sem} includes visual results of implementing the framework on Shapenet Airplane and Car classes. Figure illustrates random objects and their manipulated results by translating the latent code towards the identified latent space directions using the \(\name\). These results confirms the robustness and adaptability of \(\name\) in identifying weakly-supervised part-semantics and navigating in object level latent spaces for different object classes. 

\begin{figure}[t]
  \centering
    \includegraphics[width=1\linewidth]{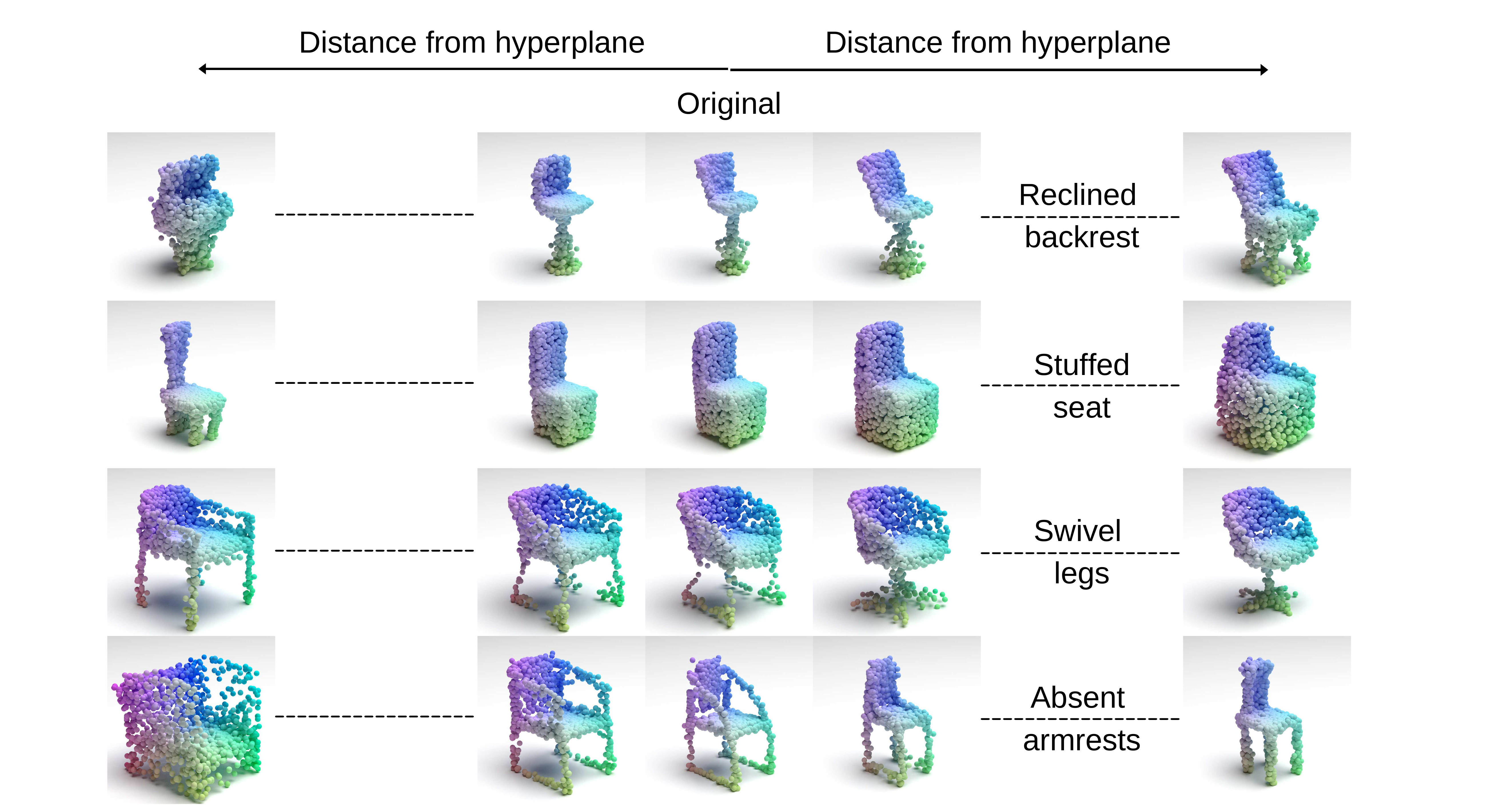}
  \caption{Controlling the extent of manipulation. The original refers to a generation close to a hyperplane. Moving the latent code farther away from the hyperplane adds unintended changes 
  , and reduce regeneration quality.
  }
  \label{fig:near_and_far_hyperplanes}
\end{figure}

\noindent
\textbf{Quality of generated and transformed objects.} 
Unlike 3D object part manipulation/editing frameworks that detach the object parts to interchange or interpolate between different objects and subsequently combine them to form the complete object, \(\name\) generates semantically edited objects from the same generator with acquired contextual knowledge of object part connections \cite{anastasia2019composite,cihan2021LPMnet}. Therefore, \(\name\) does not produce disconnected object parts or unrealistic object shapes. Since we leverage existing state-of-the-art 3D point cloud regeneration models, the edited point clouds will depict the same qualitative results of the used regeneration model. However, as shown in Fig.\ref{fig:near_and_far_hyperplanes}, if the latent code moves to the far periphery of the latent space, there can be a reduction in the regeneration quality, and an increase in the unintended changes added.

However, as shown in Fig.\ref{fig:near_and_far_hyperplanes}, if the latent code moves to the far periphery of the latent space, there can be a reduction in the regeneration quality, and an increase in the unintended changes added.

\subsection{Ablation analysis}

\begin{figure*}
  \includegraphics[width=1\linewidth]{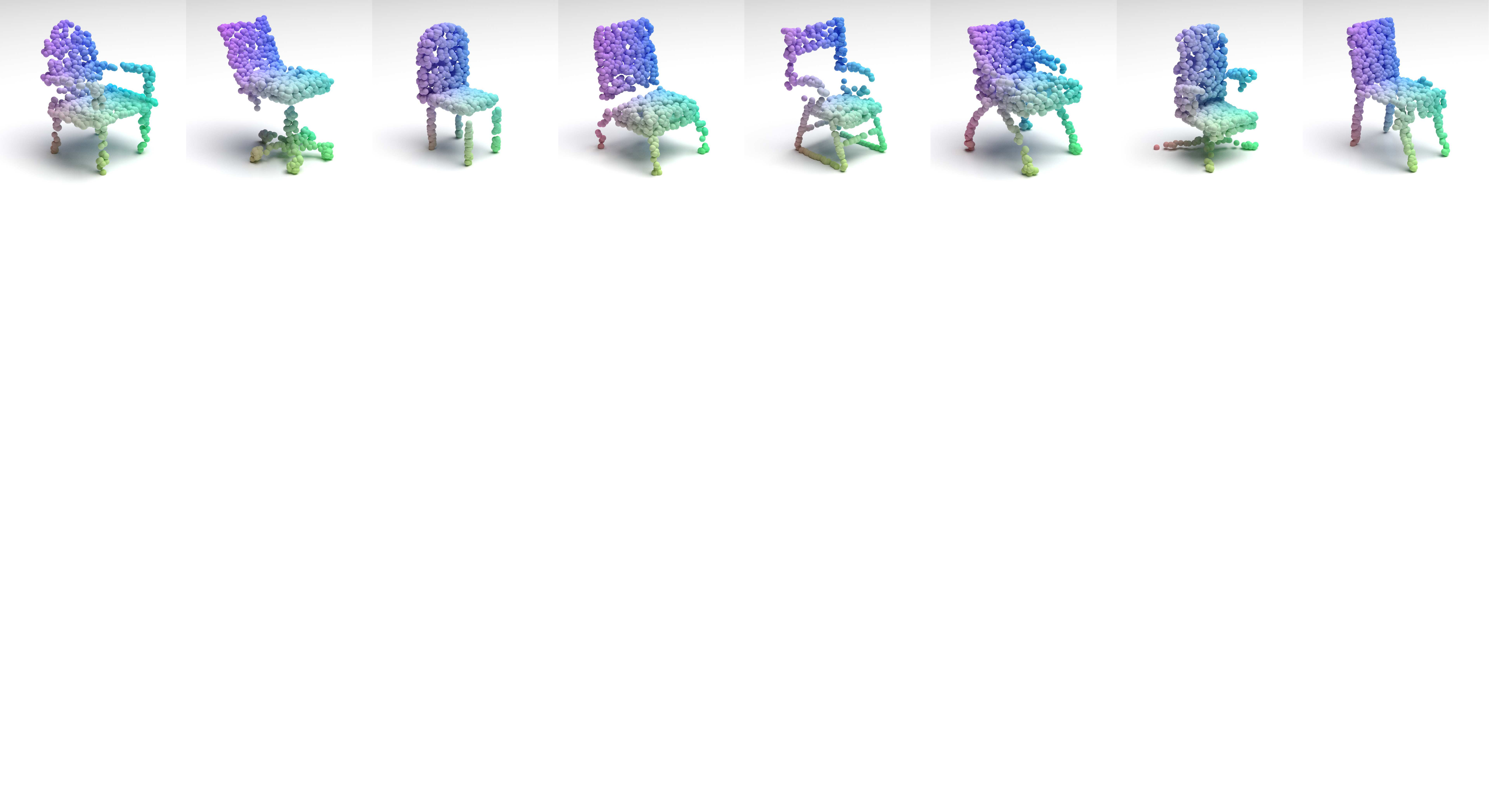}
  \caption{Examples of synthetic chairs generated by naively combining random object parts of chair class.}
  \label{fig:synthetic_examples}
\end{figure*}

\noindent \textbf{Training with synthetic data.} 
A generator trained on a certain dataset sometimes learn entangled part-level semantic features due to the entanglements introduced by the dataset. To investigate the ablation of this effect, we trained our generator on an extended ShapeNet dataset formed by introducing a set of additional examples synthesized by naively combining objects parts from different objects. The synthetic objects were created by randomly combining different segmented parts from different objects in a naive fashion. No post-processing was done to enforce a connectivity or realistic nature on these synthetic chairs. Fig.\ref{fig:synthetic_examples} shows a few examples of such synthetic chairs.

\begin{figure*}
  \includegraphics[width=1\linewidth]{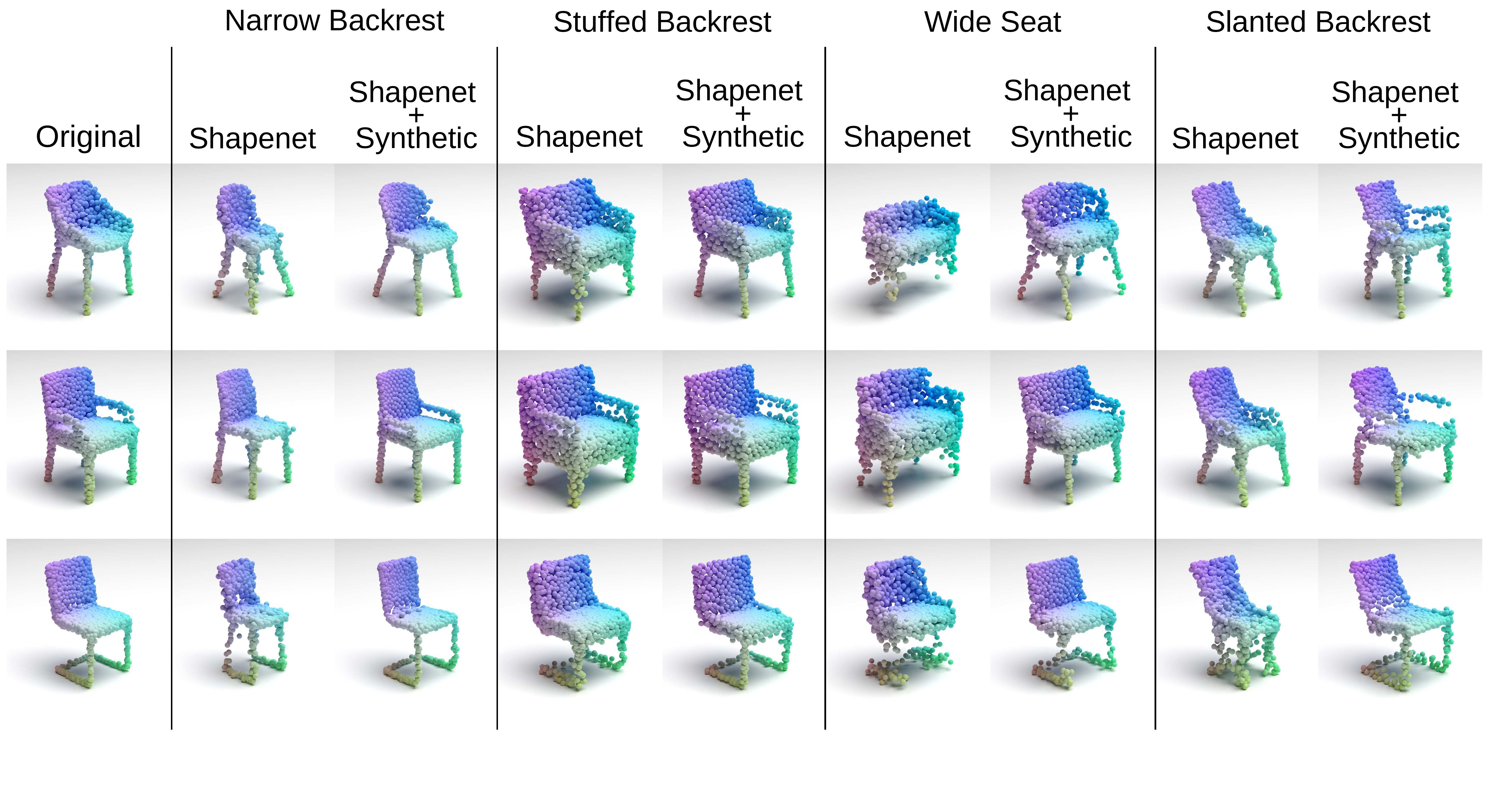}
  \caption{Manipulating the same part-level semantic in two generators trained on ShapeNet \cite{shapenet2015} dataset and Shapenet+Synthetic objects dataset respectively. Additional training on synthetic data results in further part-level semantic disentanglement, and limits the unintended changes added to the other parts.
}
  \label{fig:synthetic_train}
\end{figure*}

Nevertheless as shown in Fig.\ref{fig:synthetic_train}, we see that editing the same part-level semantics in a generator trained on this extended dataset produces objects with more disentangled part-level features, \textit{e.g. reclining the backrest without changing the seats or legs}. Moreover, exposure to diverse structures in the extended dataset enables the generator to produce objects with better quality for previously unseen feature combinations, \textit{e.g. widening the seats.}

\noindent \textbf{Using subclasses as a proxy for part semantics.}
The ShapeNet dataset provides subclass labels for its 3D objects, e.g.: `Chair' superclass has subclasses such as Tulip chair, Cantilever chair, and Lounge chair etc. 
In this ablation study, we compare the part-level semantics identified in the \(\name\) framework with Shapenet object subclasses taken as a proxy for their implied part-level semantics. \textit{E.g- Cantilever chair - for cantilevel legs, Straight chair - for straight legs, straight backrests, Armchair - for connected armrests, etc.}

We project the segmented objects parts from corresponding subclasses to the part-level latent space $L_p$ to evaluate the performance in part-level semantic identification. 
The Silhouette score calculates the ratio between mean intra-cluster distance, and the mean nearest-cluster distance. The negative values in Tab.~\ref{silhoutte_score} shows that the taking subclass annotations as a proxy for part-semantics result in overlapping clusters in the part-level latent space, indicating poor performance in distinguishing part-based semantic identification. Whereas near-zero positive values for the clusters yielded by \(\name\) indicate far superior identification of unique part-semantics.

\begin{table}[]
\centering
\scriptsize
\caption{Silhouette scores for the identified part-level latent clusters in $L_p$ and for Shapenet subclass labels taken as a proxy for part-level semantics. Part-based semantic identification using \(\name\) method deliver better clustering results than using the available subclass labels, indicating more accurate identification of unique part-level semantics.}
\begin{tabular}{lrrrr}
\toprule
 & Backrest & Seat & Legs & Armrest \\ \midrule 
 \(\name\) labels & 0.024 & 0.141 & 0.085 & 0.618\\
 ShapeNet subclass labels & -0.254 & -0.174 & -0.475 & -0.892 \\ \bottomrule
\end{tabular}
\label{silhoutte_score}
\end{table}

Next we replaced part-level semantics identified by \(\name\) with these inclusive labels of object-level subclasses and fitted SVMs in the generative latent space $L_o$  by taking each subclass label as positive examples and transformed random query latent codes normal to the identified hyperplanes. A subclass comprises of chairs with a specific combination of shape semantics in each of its constituent parts. Therefore, translating along such latent directions tend to change multiple part semantics at once, consequently decreasing the extent of controlability of the shape manipulation. Moreover, some of these subclass labels are based on the objects' functionality, and does not necessarily reflect a similarity in their shapes or structures. One such example is the \textit{Lounge chair}. The latent directions corresponding to such subclasses tend to have negligible net effect on the transformed query object. 

\begin{figure*}
  \includegraphics[width=1\linewidth]{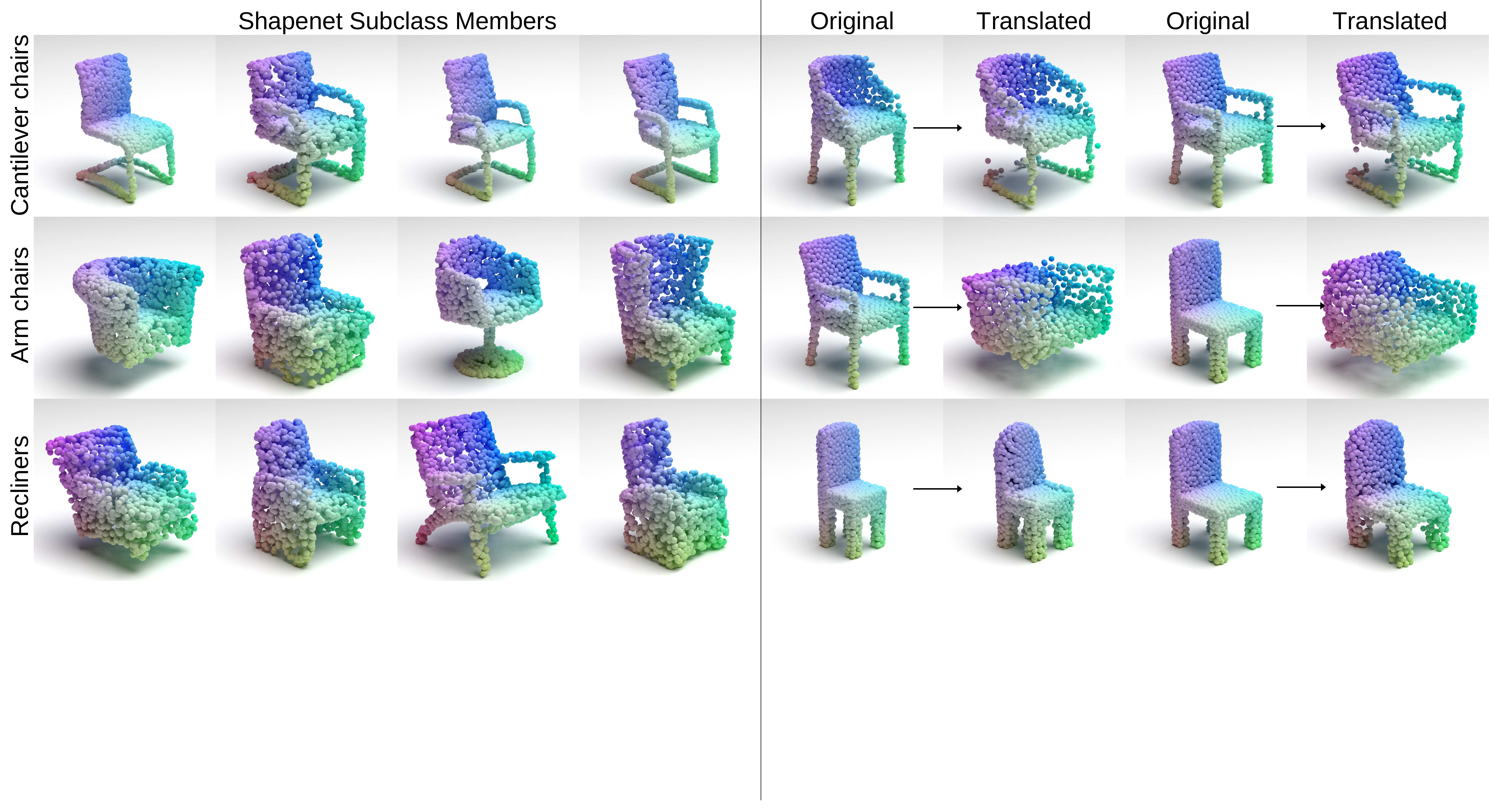}
  \caption{Left- examples of Shapenet chairs with the particular subclass label, Right- Examples of translating a random object towards the semantic direction derived using the subclass as positive examples. Row 1- (Success case) Cantilever legs added to normal chair, Row 2- (Failure case) Multiple unintended parts changed in the test chair, Row 3- (Failure case) No significant semantic change added to test chair }
  \label{fig:subclass_examples}
\end{figure*}

Fig.\ref{fig:subclass_examples} include visualizations from this ablation Study.
Row 1 shows a success case where fitting SVM and translating perpendicular to its the hyperplane for \textit{cantilever chair} subclass adds cantilever legs to a test chair. In Row 2 (Failure case), latent direction derived using \textit{Arm chair} subclass tend to change multiple part semantics of the test object at once, decreasing the controlability of the shape editing. In Row 3 (Failure case), the subclass label \textit{Recliners} is based on the objects' functionality, and does not necessarily reflect a similarity in shapes or structures. Therefore the corresponding latent direction tend to have no net effect on the transformed test object.

\begin{figure*}
  \centering
  \includegraphics[width=0.9\linewidth]{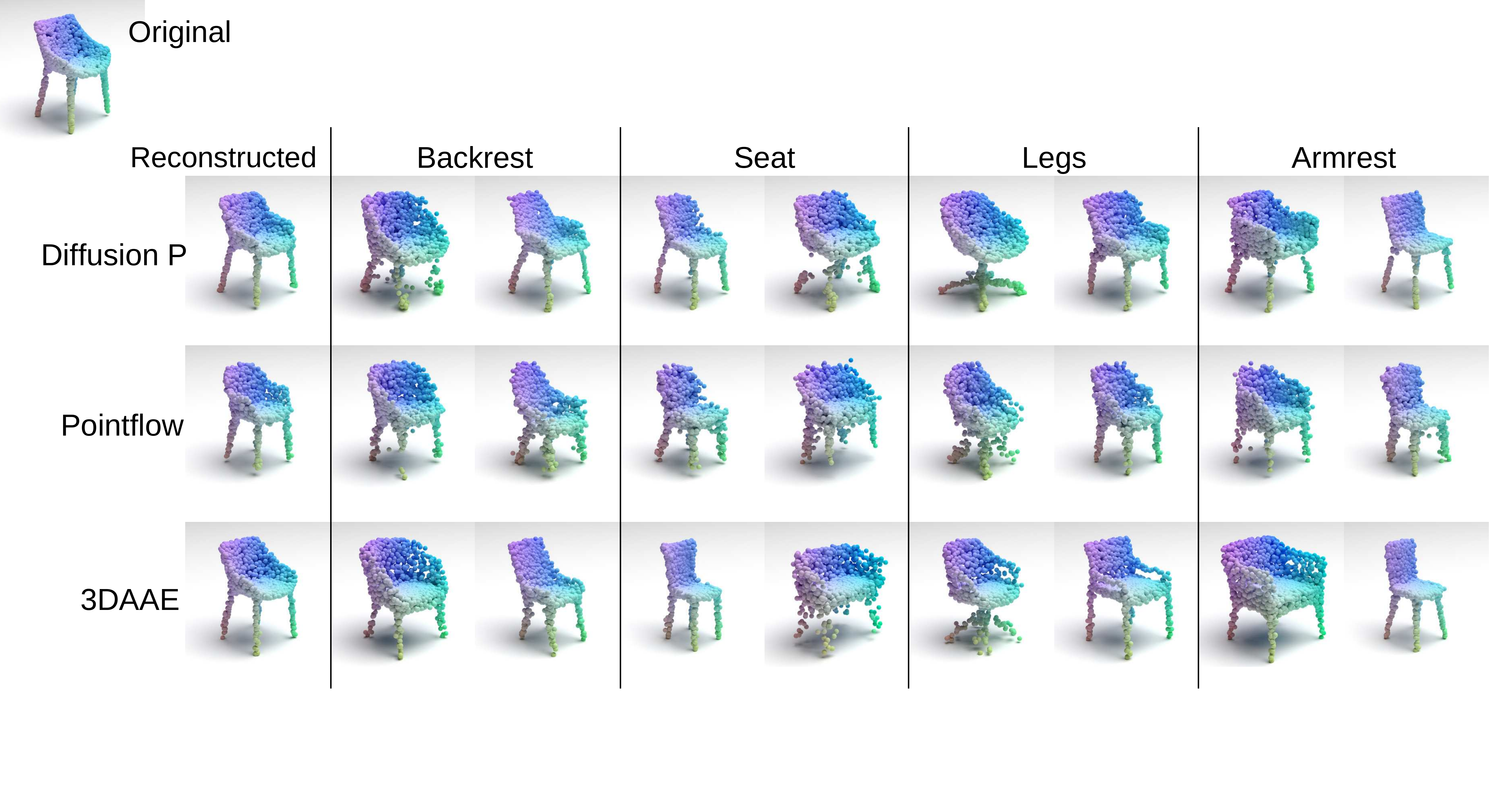}
  \caption{Implementing the \(\name\) on different pretrained 3D object generators; Diffusion Probablistic Models\cite{Shitong2021diffpm}, Pointflow\cite{guandao2020pointflow}, and 3DAAE\cite{Zamorski20203dAAE}}
  \label{fig:different generators}
\end{figure*}

\noindent \textbf{Using \(\name\) on different generative models.}
\(\name\) can be used on the generative latent spaces of any pretrained generator. Fig. \ref{fig:different generators} presents the visual results of applying our method on different generators ; Diffusion Probablistic Models\cite{Shitong2021diffpm}, Pointflow\cite{guandao2020pointflow}, and 3DAAE\cite{Zamorski20203dAAE}. We show that the proposed method can be applied on any pretrained object generative latent space, and each generative latent space may have slight differences in terms of the added changes. These changes are caused by the differences in model architectures and levels of training.
 
\bibliographystyle{splncs04}
\bibliography{egbib}

\end{document}